\documentclass{article}

\usepackage{arxiv}              % arXiv preprint style (single-column "A PREPRINT" header)

\usepackage[utf8]{inputenc}     % allow utf-8 input
\usepackage[T1]{fontenc}        % use 8-bit T1 fonts
\usepackage{microtype}          % microtypography
\usepackage{booktabs}           % professional-quality tables
\usepackage{multirow}
\usepackage{amsmath}
\usepackage{amssymb}
\usepackage{graphicx}
\usepackage{float}              % allow [H] placement for appendix tables
\usepackage{xcolor}
\usepackage{url}
\usepackage[numbers]{natbib}    % citations (\citet/\citep/\cite), numeric style
\usepackage{hyperref}           % hyperlinks
\usepackage{xspace}

\newcommand{\sysname}{ReliableTableQA}
\newcommand{\ucar}{UCAR}

\newcommand{\eg}{e.g.,\xspace}

\title{\sysname: How Much Supervision Does Reliability Annotation Need?}

\author{
  Huei-Chung Hu \\
  DOCOMO Innovations, Inc. \\
  \texttt{heidi.hu@docomoinnovations.com} \\
  \And
  Hsin-Tai Wu \\
  DOCOMO Innovations, Inc. \\
  \texttt{hwu@docomoinnovations.com} \\
  \And
  Koyo Kobayashi \\
  NTT DOCOMO, Inc. \\
  \texttt{kouyou.kobayashi.gv@nttdocomo.com} \\
}

\begin{document}

\maketitle

% ============================================================
\begin{abstract}
% Farquhar 5-sentence formula:
% 1. What we achieve  2. Why hard/important  3. How  4. Evidence  5. Best number

We introduce \sysname, a framework for training an LLM to annotate
the \emph{statistical reliability} of tabular QA results---not whether
the query is answerable, but whether the computed answer is statistically
meaningful.
In real enterprise analytics, a syntactically correct SQL query can
return a value that is too small-sample, too wide-CI, or too confounded
to act on; existing systems answer confidently in all such cases,
a failure we quantify as the \emph{Unreliable Confident Answer Rate} (\ucar).
We contribute
(1) a ten-category reliability taxonomy (R1--R10) covering hazards such as
small-sample aggregates, multiple-comparison inflation, and distribution-tail
mismatch;
(2) a program-first data pipeline that generates 50{,}000 reliability-labeled
training examples from a context-free grammar over public retail schemas,
with schema-stratified SFT/GRPO splits; and
(3) a \emph{controlled study of how much supervision calibrated reliability
annotation actually requires}.
We find that a small, schema-stratified SFT set is remarkably sufficient:
200 examples raise reliability-flag F1 from 0.61 to 0.98 and parse rate
from 0.52 to 1.00, drive \ucar\ to zero, and yield a model that generalizes
to an \emph{unseen} retail domain (Rel-F1 0.997 on held-out H\&M).
Against this strong SFT baseline, GRPO---commonly assumed essential---helps
\emph{only} when SFT is under-trained (+0.06--0.16 exact-flag-set match at
100 examples, in- and out-of-distribution) and provides no measurable benefit
once SFT is adequate, a null result we confirm across a hard compound-flag
slice, a strict exact-match metric, and out-of-distribution evaluation.
Our findings reframe reliability annotation as a \emph{data-efficiency}
problem and delineate precisely when reinforcement fine-tuning does---and
does not---pay off.
\end{abstract}

% ============================================================
\section{Introduction}

Enterprise data analytics relies heavily on tabular question answering:
a business user asks a natural-language question, an LLM generates
SQL, the query executes, and an answer is returned.
The dominant evaluation paradigm for this pipeline measures
\emph{execution accuracy}---whether the SQL returned the right value
\cite{yu2018spider,li2023bird}.
This framing treats any executed result as equally trustworthy.

It is not.
Consider a query computing the average smartphone usage score for
customers in a prefecture with $n=3$ matching rows, a top-prefecture
ranking that reverses under Bonferroni correction across 38 simultaneous
comparisons, or an aggregate that silently excludes 15\% of rows due to
null values.
In all three cases, the SQL is correct: it executed, returned a value, and
that value is an accurate aggregate of the matching rows.
Yet none of these results is safe to act on.
Current systems---both zero-shot LLMs and fine-tuned text-to-SQL
models---report all three with equal confidence.

We formalize this failure as the gap between \emph{answerability} and
\emph{reliability}. Prior work on LLM abstention addresses the former:
should the model refuse because the data does not exist, the question has
no factual answer, or the SQL cannot be written
\cite{kirichenko2025abstentionbench,lee2024trustsql,chen2025rts}?
We address the orthogonal axis: the SQL ran successfully, the data exists,
but \emph{the statistical properties of the result make it misleading}.
We call this mode \emph{answerable-but-unreliable} and introduce
the \emph{Unreliable Confident Answer Rate} (\ucar) to measure it.

\paragraph{Contributions.}
\begin{itemize}
  \item \textbf{Taxonomy.} Ten reliability hazards (R1--R10) covering
    small-sample aggregates, confidence-interval width, multiple-comparison
    inflation, missing-data bias, subgroup imbalance, single-observation
    dominance, Simpson-reversal risk, distribution-tail mismatch,
    rare-event proportions, and cross-tab sparseness.
    Each hazard is programmatically detectable from the executed query result.

  \item \textbf{Pipeline.} A program-first data generation method that
    inverts the standard LLM-paraphrase approach: a context-free grammar
    generates structurally diverse SQL queries over real retail schemas,
    a deterministic profiler assigns multi-label reliability annotations,
    and a 3B LLM realizes natural-language questions on top with
    embedding-distance diversity filtering.
    This avoids the mode-collapse failure of naive LLM-driven generation;
    Appendix~\ref{app:pipeline} gives the grammar, the profiler rules, and a
    worked example.

  \item \textbf{When does RL help?} A controlled SFT-size ablation showing
    that calibrated reliability annotation is largely a \emph{data-efficiency}
    problem. A model that jointly generates a SQL answer \emph{and} a structured
    reliability assessment is trained with SFT and then GRPO, whose rewards are
    tied directly to DuckDB-executable ground truth. We find a few hundred
    schema-stratified SFT examples suffice and transfer to an unseen domain,
    while GRPO helps \emph{only} when the SFT base is under-trained and adds
    nothing once it is adequate---robust to hard slices, a strict exact-match
    metric, and out-of-distribution evaluation.
\end{itemize}

\begin{figure*}[t]
  \centering
  \includegraphics[width=\linewidth]{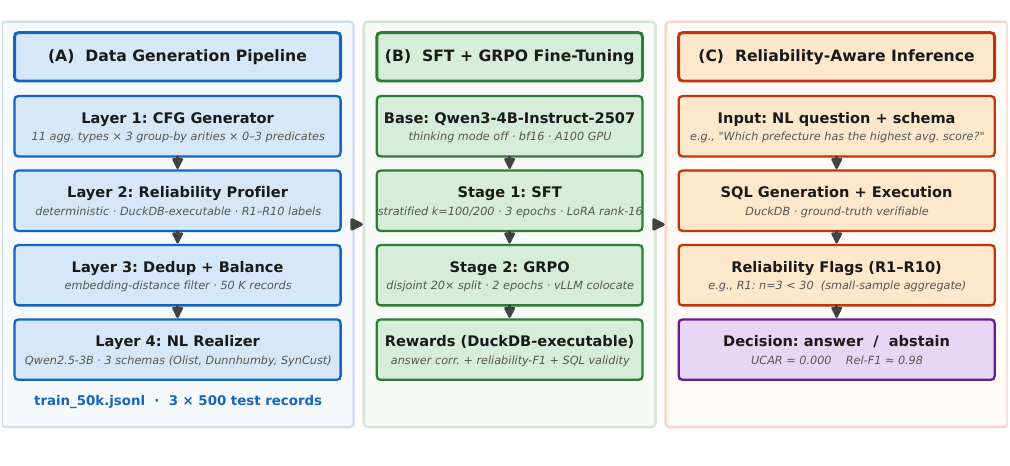}
  \caption{%
    \textbf{Overview of the \sysname\ framework.}
    \textbf{(A)~Data generation:} a context-free grammar (Layer~1) emits
    structurally diverse SQL queries, a deterministic profiler (Layer~2)
    assigns R1--R10 reliability labels, deduplication and balancing
    (Layer~3) produce a 50\,K labeled pool, and a small LLM (Layer~4)
    realizes natural-language questions. SynCust is an abbreviation of Synthetic Customer.
    \textbf{(B)~Fine-tuning:} Qwen3-4B-Instruct is SFT-trained on a small
    schema-stratified subset (we sweep $k$ from 50 to 1000; GRPO is run at
    $k{=}100$ and $200$), then GRPO-trained
    on a disjoint split ($20\times$ the SFT size) with three
    DuckDB-executable reward signals and no reward model.
    \textbf{(C)~Inference:} the fine-tuned model generates SQL, executes it,
    raises applicable reliability flags (R1--R10), and decides whether to
    answer or abstain.
    Our data-efficient SFT model achieves \ucar\,=\,0.000 and
    Rel-F1\,$\approx$\,0.98 across schemas (0.997 on the held-out H\&M domain),
    with GRPO adding measurable benefit only when SFT is under-trained. 
  }
  \label{fig:overview}
\end{figure*}

Experiments across one synthetic and two public retail schemas (Synthetic Customer,
Olist Brazilian e-commerce, Dunnhumby household panel), a held-out fourth
domain (H\&M), and harder evaluation slices show that a data-efficient SFT
model drives unreliable confident answers to zero (\ucar\ = 0.000) and reaches
0.98 reliability-flag F1, while zero-shot and prompt-only baselines fail at
rates of 5.5\% and 2.3\% respectively---and that GRPO's contribution is
confined to the under-trained regime.
Figure~\ref{fig:overview} illustrates all three components of the \sysname\ framework.

% ============================================================
\section{Problem Formulation}
\label{sec:problem}

\paragraph{Setting.}
Let $\mathcal{T}$ be a relational table and $q$ a natural-language question.
A tabular QA system produces a response $r = (d, s, \mathbf{f}, e)$, where:
$d \in \{\texttt{answer}, \texttt{abstain}\}$ is a decision,
$s$ is a SQL query,
$\mathbf{f} \subseteq \{R1,\ldots,R10\}$ is a set of reliability flags, and
$e$ is a natural-language reliability explanation.

\paragraph{Reliability profiler.}
Given $s$ and $\mathcal{T}$, a deterministic profiler $\phi$ executes $s$
and computes a binary reliability vector
$\phi(s, \mathcal{T}) \in \{0,1\}^{10}$,
one entry per hazard (Table~\ref{tab:taxonomy}).
A result is \emph{reliable} ($\mathbf{f} = \emptyset$) if no hazard fires;
otherwise it is \emph{unreliable} with label set $L = \{R_i : \phi_i = 1\}$.

\paragraph{Evaluation metrics.}
We assess systems on five axes:
\begin{itemize}
  \item \textbf{Rel-F1}: Micro-F1 of predicted flags $\mathbf{f}$ against ground-truth
    labels $L$, averaged across records.
  \item \textbf{\ucar}: Among records with $L \neq \emptyset$, the fraction
    where the model outputs $d=\texttt{answer}$, $\mathbf{f}=\emptyset$,
    and \texttt{confidence=high}. Lower is better; 0.0 means the model
    never confidently answers an unreliable query without flagging it.
  \item \textbf{Answer accuracy}: Whether the model's SQL, when executed,
    produces a result matching the gold SQL's output.
  \item \textbf{SQL validity}: Whether the model's SQL executes without error.
  \item \textbf{Abstention accuracy}: Whether the model's decision
    $d$ matches the ground-truth abstain/answer label.
    The ground-truth is \texttt{abstain} only when R1 fires at the hard threshold ($n < 5$);
    queries with $5 \leq n < 30$ raise R1 as a warning but retain \texttt{answer}
    as the ground-truth decision.
\end{itemize}

% ============================================================
\section{Reliability Taxonomy}
\label{sec:taxonomy}

Table~\ref{tab:taxonomy} defines the ten reliability hazards.
Each hazard is detectable from the query result without access to external
knowledge: only the executed rows, group sizes, and column statistics are
required.
This is by design---the system must operate as a post-execution annotator,
not a pre-execution filter.

\begin{table*}[t]
\centering
\small
\begin{tabular}{llp{7.5cm}}
\toprule
\textbf{ID} & \textbf{Hazard} & \textbf{Detection criterion} \\
\midrule
R1 & Small-sample aggregate
   & Any result group has $n < 30$ (warn) or $n < 5$ (abstain) \\
R2 & Wide confidence interval
   & Bootstrap 95\% CI width $> 0.5 \times |\hat{\mu}|$ \\
R3 & Multiple-comparison inflation
   & Top-$k$ ranking query; Bonferroni or BH-FDR correction eliminates winner \\
R4 & Missing-data bias
   & NaN exclusion rate $> 10\%$ of rows for any involved column \\
R5 & Subgroup imbalance
   & $\min(\text{group sizes}) / \max(\text{group sizes}) < 0.1$ or both $< 10$ \\
R6 & Single-observation driver
   & Removing 1--2 maximum-value rows changes the top-$k$ ranking \\
R7 & Simpson-reversal candidate
   & Group-level direction reverses under stratification by an available covariate \\
R8 & Distribution-tail mismatch
   & Skewness $> 2$ or mean/median $> 1.5$ for a mean-based aggregate \\
R9 & Rare-event proportion
   & Proportion query with numerator $< 30$ (Wilson CI dominates) \\
R10 & Categorical sparseness
   & Cross-tab with expected cell count $< 5$ (chi-squared test invalid) \\
\bottomrule
\end{tabular}
\caption{The ten reliability hazards (R1--R10). Each is computed deterministically
  from the executed query result. A query can trigger multiple hazards simultaneously.}
\label{tab:taxonomy}
\end{table*}

The taxonomy is grounded in classical statistical practice---these failure modes
appear in textbook treatments of small-sample inference, multiple testing, and
exploratory data analysis \cite{gelman2013bda,tukey1977eda}.
We formalize them as a \emph{learnable, evaluatable label space} for tabular QA:
a model can be trained to predict $\phi(s, \mathcal{T})$ from the user's question
and the profiler statistics, and evaluated by F1 against ground truth.

% ============================================================
\section{Data Pipeline}
\label{sec:pipeline}

Constructing a reliability-labeled training set requires queries that
\emph{densely cover} all ten hazard categories, including rare ones such as R3
(multiple-comparison inflation) and R8 (distribution-tail mismatch).
Naive LLM-driven question generation fails at this task: prompted to produce
diverse analytics questions, models collapse to high-frequency templates
(\eg, ``how many customers in state X?'') that over-represent common patterns
and under-represent rare structural hazards such as R3 and R8.
We invert the dependency (Appendix~\ref{app:pipeline} details the grammar, the
deterministic profiler rules, and a worked end-to-end example).

\paragraph{Layer 1: CFG-based SQL generation.}
A context-free grammar over each schema generates SQL queries with
combinatorial structural coverage across aggregate type
(AVG, MEDIAN, SUM, MAX, MIN, STDDEV, P90, P10, COUNT, COUNT DISTINCT, RATIO),
group-by arity (0, 1, 2),
predicate count (0--3), and
output cardinality bucket (1, 2--10, 11--50, 51+).
Literal values (\eg, categorical filter values, numeric thresholds) are sampled
from the actual column distributions in each table.
We discard queries with zero-cardinality results or trivially-constant outputs.

\paragraph{Layer 2: Reliability profiler.}
Each query is executed against its table via DuckDB.
A deterministic profiler computes the full R1--R10 signal vector from
the result set and assigns multi-label annotations.
The profiler is implemented in pure Python and requires no external APIs
(bootstrap CI uses $B=1000$ resamples; BH-FDR uses the Benjamini-Hochberg
procedure~\cite{benjamini1995controlling}).

\paragraph{Layer 3: Deduplication and balancing.}
We collapse queries to structural templates by abstracting numeric and string
literals, then sample one representative per template.
Labels with low natural occurrence (R3, R8) are oversampled to a minimum of
3{,}000 instances per label before NL generation.

\paragraph{Layer 4: NL question realization.}
A 3B instruction-tuned LLM (Qwen2.5-3B-Instruct~\cite{qwen25_2025}) generates two natural-language
question variants per query using five persona types
(business analyst, data scientist, product manager, executive, technical operator).
A greedy embedding-distance filter (all-MiniLM-L6-v2) removes variants too
similar to previously accepted questions within each batch ($\cos > 0.85$).

\paragraph{Layer 5: Combination and split.}
We generate 30{,}000 queries per schema (Synthetic Customer, Olist, Dunnhumby),
apply NL generation, then sample 50{,}000 records stratified by schema into
the training set.
Three test sets of 500 records each are held out before NL generation
and evaluated with template-based question substitutes.
Table~\ref{tab:dataset} summarizes label distributions.

\begin{table}[t]
\centering
\small
\begin{tabular}{lrrr}
\toprule
\textbf{Label} & \textbf{SynCust} & \textbf{Olist} & \textbf{Dunnhumby} \\
\midrule
R1 & 71.0\% & 57.0\% & 76.0\% \\
R2 &  5.3\% &  2.6\% & 10.7\% \\
R3 &  8.5\% &  9.0\% & 12.4\% \\
R4 & 14.0\% &  0.0\% &  0.0\% \\
R5 & 44.0\% & 57.0\% & 54.0\% \\
R6 &  5.5\% &  4.5\% & 11.6\% \\
R7 &  9.5\% &  3.5\% & 11.5\% \\
R8 &  0.1\% & 32.0\% & 19.0\% \\
R9 &  8.5\% &  5.0\% & 12.6\% \\
R10 & 9.0\% &  4.5\% & 13.0\% \\
Reliable & 12.0\% & 15.0\% & 10.0\% \\
\bottomrule
\end{tabular}
\caption{Reliability label distributions per schema in the training set (50k total).
  Labels are multi-hot; rows do not sum to 100\%.
  R4 is absent from Olist and Dunnhumby because their flattened tables
  have no missing values in the joined columns.
  R8 is naturally rich in the retail schemas due to heavy-tailed spend distributions. SynCust is an abbreviation for Synthetic Customer.}
\label{tab:dataset}
\end{table}

% ============================================================
\section{Model}
\label{sec:model}

\subsection{Output Format}

The model produces a structured JSON response containing the following fields:
\texttt{decision} (answer/abstain),
\texttt{sql} (executable DuckDB SQL),
\texttt{result} (one-sentence summary),
\texttt{reliability\_flags} (subset of R1--R10 flag names),
\texttt{reliability\_explanation} (one--two sentence explanation citing specific statistics),
\texttt{confidence} (high/medium/low), and
\texttt{evidence} (key numeric values from the profiler).

The model receives the table description, natural-language question, and
the full profiler statistics block (group sizes, CI width, null rates, etc.)
as input context.
The model annotates reliability \emph{given} the profiler statistics,
not by recomputing them from raw data.

\subsection{Stage 1: Supervised Fine-Tuning}

We fine-tune Qwen3-4B-Instruct-2507 \cite{qwen3_2025} with LoRA
(rank 16, $\alpha=32$, all projection modules) using TRL's SFTTrainer
\cite{vonwerra2020trl}.
The training objective is next-token prediction on the assistant turn only
(response-masked loss).
Rather than train on the full 50k pool, we draw a small \emph{schema-stratified}
subset of size $k$ (balanced across the ten hazards and the schemas) and train
for 3 epochs at effective batch size 16 (micro-batch 2, gradient accumulation 8).
The large, diverse pool is precisely what makes this stratification possible---
every (hazard, schema) cell, including rare ones such as R3 and R8, has enough
candidates to sample a balanced subset---so the data-efficiency result below is a
finding about how training data is \emph{selected}, not an argument against
generating it.
We report $k{=}100$ (S1) and $k{=}200$ (S2), reaching final train loss
$1.16$ and $0.68$ respectively. As Section~\ref{sec:experiments} shows, the
larger set is what lets the model reliably emit its structured output format;
the SFT stage teaches both the format and the deterministic mapping from
profiler statistics to reliability flags.

\subsection{Stage 2: GRPO Fine-Tuning}

On top of the merged SFT weights we apply Group Relative Policy Optimization
(GRPO) \cite{shao2024deepseekmath} and study its contribution as a function of
SFT adequacy. We use three deterministic, DuckDB-executable reward functions
(no learned reward model); all are pre-weighted and summed by TRL:

\begin{itemize}
  \item \textbf{Answer correctness} ($+0.30$): the generated SQL is executed
    against the table and its result compared to the gold SQL output using
    exact match or 1\% relative tolerance for scalars.
  \item \textbf{Reliability-label F1} ($+0.25$): micro-F1 of the predicted
    R1--R10 flag set against the profiler ground truth.
  \item \textbf{SQL execution validity} ($+0.15$): the generated SQL
    executes without error.
\end{itemize}

We train with full-parameter fine-tuning (SFT LoRA merged into base weights,
no new adapter) for 2 epochs on the \emph{GRPO split}---disjoint from the SFT
set and $20\times$ its size (2{,}000 records at S1, 5{,}000 at S2)---with
$G{=}4$ generations per prompt, micro-batch 1, gradient accumulation 8, peak LR
$5\times10^{-6}$, sampling temperature $0.8$, and KL coefficient $\beta{=}0.04$.
The SFT checkpoint initializes both the policy and (a frozen copy as) the KL
reference. We use vLLM colocate generation (${\sim}9{\times}$ speedup over
standard HF generation); because Qwen3 disables its KV cache under gradient
checkpointing (raising per-step time from ${\sim}9$s to ${\sim}160$s), we set
\texttt{gradient\_checkpointing=False}. Because the adequate-SFT base already
nears the reliability-F1 and parse-rate ceilings, GRPO's advantage signal there is
small---consistent with its negligible effect at S2 (Section~\ref{sec:grpo_when}).

% ============================================================
\section{Experiments}
\label{sec:experiments}

\subsection{Setup}

Our central question is \emph{how much supervision calibrated reliability
annotation requires}, and whether GRPO adds value beyond a well-trained SFT
base. We therefore evaluate two non-fine-tuned baselines and an
\emph{SFT-size sweep} in which both the SFT-only and the SFT+GRPO checkpoint
are evaluated at each setting:

\begin{enumerate}
  \item \textbf{Zero-shot}: Qwen3-4B-Instruct-2507 with a system prompt
    that specifies the JSON output format and lists flag names but provides
    no guidance on when to raise them.
  \item \textbf{Prompt-only}: Same base model with the full reliability-aware
    system prompt, which describes all ten hazards and their thresholds.
  \item \textbf{SFT($k$) / SFT+GRPO($k$)}: models fine-tuned on a
    schema-stratified SFT set of size $k$, then GRPO-trained on a disjoint set
    ($\!\times\,20$). We report $k{=}100$ (\textbf{S1}) and $k{=}200$ (\textbf{S2}),
    each evaluated both before and after GRPO (LoRA merged for SFT;
    full-FT for GRPO).
\end{enumerate}

All models use greedy decoding (\texttt{do\_sample=False}) and are scored
against DuckDB-executed ground truth. We evaluate on three in-distribution
schemas (Synthetic Customer, Olist, Dunnhumby; 500 records each), and to stress-test the
findings, on (i)~a \emph{hard slice} of compound-hazard records
($\geq 3$ gold flags, $n{=}349$) scored with strict exact-flag-set match, and
(ii)~a \emph{held-out fourth domain} (H\&M fashion retail; 500 records), never
seen in training. SFT runs complete in well under an hour; GRPO runs take
5--14h on a single A100 80GB.

\subsection{SFT is Data-Efficient---Past a Threshold}

Table~\ref{tab:main_results} reports in-distribution results (means across three
schemas) for the baselines and the SFT-size sweep.

\begin{table*}[t]
\centering
\small
\begin{tabular}{lrrrrr}
\toprule
\textbf{Model} & \textbf{Rel-F1} $\uparrow$ & \textbf{\ucar} $\downarrow$
  & \textbf{Ans. Acc.} $\uparrow$ & \textbf{Abst. Acc.} $\uparrow$
  & \textbf{Parse} $\uparrow$ \\
\midrule
\multicolumn{6}{l}{\textit{No fine-tuning}} \\
Zero-shot              & 0.817 & 0.055 & 0.659 & 0.649 & 0.998 \\
Prompt-only            & 0.804 & 0.023 & 0.587 & 0.866 & 0.989 \\
\midrule
\multicolumn{6}{l}{\textit{SFT-size sweep (SFT-only; \texttt{+GRPO} where run)}} \\
SFT(50)                & 0.387 & 0.000 & 0.089 & 0.607 & 0.153 \\
SFT(100)               & 0.608 & 0.000 & 0.350 & 0.752 & 0.521 \\
\quad +GRPO            & 0.667 & 0.000 & 0.435 & 0.757 & 0.601 \\
SFT(150)               & 0.940 & 0.001 & 0.721 & 0.977 & 0.999 \\
SFT(200)               & 0.978 & 0.001 & 0.723 & 0.978 & 1.000 \\
\quad +GRPO            & 0.978 & 0.000 & 0.719 & 0.978 & 1.000 \\
SFT(500)               & 0.996 & 0.000 & 0.720 & 0.976 & 0.996 \\
SFT(1000)              & 1.000 & 0.000 & 0.730 & 0.979 & 1.000 \\
\bottomrule
\end{tabular}
\caption{In-distribution results, means across Synthetic Customer, Olist, and Dunnhumby
  (500 records each). \ucar\ = Unreliable Confident Answer Rate. \emph{Parse} =
  fraction of outputs that parse to valid JSON with executable SQL.
  The SFT-only sweep ($k\in\{50,\dots,1000\}$) brackets a sharp data-efficiency
  threshold: parse rate collapses below $k{=}100$ ($0.15$ at $k{=}50$,
  $0.52$ at $k{=}100$), saturates by $k{=}150$--$200$ (Rel-F1 $0.94$--$0.98$,
  parse rate $\approx\!1.0$), and is flat out to $k{=}1000$---an order of magnitude
  more data adds nothing. \texttt{+GRPO} (run at $k{=}100,200$) partially
  recovers the under-fit S1 deficit but is a no-op at S2. See
  Figure~\ref{fig:sft-curve}.}
\label{tab:main_results}
\end{table*}

\begin{figure}[t]
  \centering
  \includegraphics[width=0.55\linewidth]{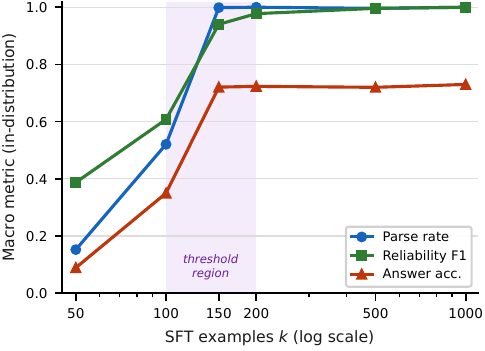}
  \caption{%
    \textbf{Reliability annotation is data-efficient: a threshold, not a
    scaling, phenomenon.}
    SFT-only (no GRPO) performance versus the number of schema-stratified SFT
    examples $k$ (log scale), macro-averaged over the three in-distribution
    schemas (Synthetic Customer, Olist, Dunnhumby; 500 records each).
    All three metrics---output \emph{parse rate} (valid JSON\,/\,SQL),
    reliability-F1, and answer accuracy---rise steeply across the shaded
    \textbf{threshold region}: below it the model is under-trained and emits
    malformed output (parse rate $0.15$ at $k{=}50$ and $0.52$ at $k{=}100$),
    which floors every metric. By $k{=}150$ the parse rate is already $0.999$,
    and the curve then saturates---$k{=}200$, $500$, and $1000$ form a
    \emph{plateau} (parse $\geq0.996$, Rel-F1 $0.98$--$1.00$), so an order of
    magnitude more SFT data buys no further gain. The knee thus falls
    \emph{between $k{=}100$ and $150$}: roughly $150$ stratified examples
    suffice for calibrated reliability annotation. This curve is SFT-only by
    design; GRPO (not shown) helps only in the under-trained regime left of the
    threshold and adds nothing once SFT is adequate
    (Table~\ref{tab:grpo_delta}).%
  }
  \label{fig:sft-curve}
\end{figure}

\paragraph{A sharp data-efficiency threshold between 100 and 200 examples.}
At S2 (200 stratified examples) the SFT model already reaches Rel-F1 $0.978$,
answer accuracy $0.723$, parse rate $1.000$, and \ucar\ $0.000$---beating
the strongest non-fine-tuned baseline (zero-shot) on every axis. At S1 (100
examples) the same recipe is \emph{under-trained}: parse rate collapses to
$0.521$ (roughly a third of generations are malformed JSON/SQL), dragging
Rel-F1 to $0.608$ and answer accuracy to $0.350$---below zero-shot. The failure
at S1 is one of \emph{output form}, not reliability reasoning: among records
that parse, answer accuracy is $0.65$--$0.80$, at or above the baselines. Thus a
few hundred well-chosen examples are sufficient, but a few hundred are also
\emph{necessary}---below the threshold the model cannot reliably emit its own
output format.
Figure~\ref{fig:sft-curve} plots the full SFT-size curve
$k\in\{50,100,150,200,500,1000\}$: the transition is sharp, with the knee
between $k{=}100$ and $150$ (parse rate $0.52\!\to\!0.999$), and $k{=}200$
onward forms a flat plateau---an order of magnitude more data ($k{=}1000$) adds
nothing further.

\paragraph{Fine-tuning eliminates unreliable confident answers.}
Both SFT settings drive \ucar\ to $\approx\!0.000$---the model essentially never
asserts a high-confidence answer on an unreliable result without flagging
it---versus $5.5\%$ (zero-shot) and $2.3\%$ (prompt-only). This behavior is
instilled by SFT alone and is preserved by GRPO.

\paragraph{Prompt-only vs.\ zero-shot trade-off.}
Adding the detailed reliability system prompt improves \ucar\ ($0.055\!\to\!0.023$)
and abstention accuracy ($0.649\!\to\!0.866$) but \emph{reduces} answer accuracy
($0.659\!\to\!0.587$) and SQL validity ($0.997\!\to\!0.945$): the verbose
instructions crowd out SQL-generation capacity, a trade-off fine-tuning resolves.

\subsection{When Does GRPO Help?}
\label{sec:grpo_when}

Table~\ref{tab:main_results} already hints at the answer: GRPO lifts the
under-trained S1 model substantially (Rel-F1 $+0.059$, answer accuracy $+0.085$,
parse rate $+0.081$) but is a no-op at S2. To test whether the S2 null
result is merely a ceiling artifact of saturated metrics, we re-score with a
strict \emph{exact-flag-set match} (the prediction must equal the gold flag set
exactly) on three increasingly demanding regimes. Table~\ref{tab:grpo_delta}
reports the GRPO contribution (SFT+GRPO $-$ SFT) at each setting.

\begin{table}[t]
\centering
\footnotesize
\setlength{\tabcolsep}{4pt}
\begin{tabular}{lrr}
\toprule
\textbf{Evaluation regime} & \textbf{S1 $\Delta$} & \textbf{S2 $\Delta$} \\
 & \textit{(100 ex)} & \textit{(200 ex)} \\
\midrule
In-distribution (full, $n{=}1500$)   & $+0.111$ & $+0.009$ \\
Hard slice ($\geq 3$ flags, $n{=}349$) & $\mathbf{+0.158}$ & $+0.000$ \\
Out-of-distribution (H\&M, $n{=}500$) & $+0.058$ & $-0.004$ \\
\bottomrule
\end{tabular}
\caption{GRPO contribution (exact-flag-set-match $\Delta$ = SFT+GRPO $-$ SFT) at
  each SFT size, across three evaluation regimes. GRPO's benefit is concentrated
  in the under-trained regime (S1) and \emph{grows} on the hard slice
  ($+0.158$), confirming the slice is discriminative; at S2 it is flat-to-negative
  everywhere, including out-of-distribution.}
\label{tab:grpo_delta}
\end{table}

\paragraph{GRPO is a low-data crutch, not an independent capability.}
Two facts establish this. First, the hard slice is \emph{discriminative}: it
amplifies GRPO's S1 benefit from $+0.111$ to $+0.158$ exact-match, so it can
detect a real GRPO effect when one exists. Second, on that same slice the S2
contribution is exactly $+0.000$, and remains $\approx\!0$ out-of-distribution
($-0.004$). The S2 null is \emph{not} a ceiling artifact: answer
accuracy on the hard slice has ample headroom yet GRPO does not improve it
($0.567$ SFT vs.\ $0.547$ SFT+GRPO), and on the hardest records ($\geq 4$ flags)
GRPO mildly \emph{hurts} answer accuracy ($0.720\!\to\!0.640$). Across four
independent angles---the standard test, a compound-hazard slice, a strict
metric, and a held-out domain---the conclusion is consistent: \textbf{GRPO with
executable rewards rescues an under-trained SFT base but adds nothing, and can
slightly hurt, once the base is adequate.}

\subsection{Domain Generalization, Including Out-of-Distribution}

A natural worry is that the data-efficient S2 model merely memorized the three
training schemas' column vocabularies. It did not. Table~\ref{tab:generalization}
shows the S2 SFT model across all three in-distribution schemas \emph{and} a
held-out fourth domain---H\&M fashion retail, whose schema, columns, and value
distributions were never seen in training.

\begin{table}[t]
\centering
\footnotesize
\setlength{\tabcolsep}{4pt}
\begin{tabular}{lrrr}
\toprule
\textbf{Domain} & \textbf{Rel-F1} & \textbf{Ans. Acc.} & \textbf{Parse} \\
\midrule
SynCust        & 0.983 & 0.788 & 1.000 \\
Olist         & 0.966 & 0.662 & 1.000 \\
Dunnhumby     & 0.984 & 0.720 & 1.000 \\
\midrule
H\&M \textit{(held-out, OOD)} & \textbf{0.997} & 0.760 & 1.000 \\
\bottomrule
\end{tabular}
\caption{The S2 SFT(200) model generalizes across domains, including a
  \emph{held-out} fourth domain (H\&M) never seen in training: Rel-F1
  $0.97$--$1.00$ and perfect parse rate throughout. The under-trained S1
  model, by contrast, degrades out-of-distribution (parse rate
  $0.521\!\to\!0.386$), confirming that adequate SFT---not GRPO---is what buys
  robust generalization. (H\&M lacks R4/R8 by construction, so its absolute
  scores are not a \emph{harder} test; the point is that domain transfer
  does not collapse.) SynCust is an abbreviation of Synthetic Customer}
\label{tab:generalization}
\end{table}

The S2 model holds Rel-F1 $0.97$--$1.00$ and perfect parse rate on a domain
it never trained on, indicating it learned \emph{general} reliability reasoning
rather than schema-specific pattern matching. The under-trained S1 model instead
degrades out-of-distribution (parse rate $0.521\!\to\!0.386$): robustness comes
from adequate SFT, and---consistent with Section~\ref{sec:grpo_when}---GRPO adds
nothing to it OOD ($\Delta = -0.004$).

% ============================================================
\section{Related Work}
\label{sec:related}

\paragraph{LLM abstention and answerability.}
AbstentionBench \cite{kirichenko2025abstentionbench} introduces
a large-scale benchmark for LLM abstention across four answerability categories
(unknown answers, false premises, subjective questions, outdated information),
finding that reasoning fine-tuning degrades abstention by 24\%.
TrustSQL \cite{lee2024trustsql} and RTS
\cite{chen2025rts} address abstention at the
SQL-generation layer---refusing to generate SQL when the schema does not support
the question.
Our work addresses an orthogonal axis: abstaining or flagging after the query
executes, because the \emph{result itself} is statistically unreliable.
A correctly-generated, successfully-executed SQL can still trigger R1--R10.

\paragraph{Calibration and uncertainty in text-to-SQL.}
\citet{liu2025calibrating} calibrate model confidence in SQL
generation using sub-clause frequency signals, targeting SQL \emph{correctness}
rather than result reliability.
Node-level uncertainty estimation \cite{hasson2025nodeue} estimates
per-AST-node uncertainty about structural correctness.
In contrast, our calibration signal is data-external: bootstrap confidence
intervals, sample sizes, and multiple-testing corrections applied to the
\emph{executed result}---independent of model internals.

\paragraph{GRPO for reasoning.}
DeepSeek-Math \cite{shao2024deepseekmath} introduced GRPO as an
efficient alternative to PPO for training mathematical reasoning.
GRPO has since been applied to code generation and SQL verification
\cite{deepseekr1_2025,pourreza2025reasoningsql}.
We apply GRPO with fully executable, DuckDB-backed reward functions,
which sidesteps learned reward model error entirely.
Whereas these works report \emph{unconditional} gains from reinforcement learning
on SQL and reasoning tasks, our controlled SFT-size sweep shows the benefit is
\emph{contingent on SFT adequacy}: substantial when the supervised base is
under-trained but negligible once it is sufficient (Section~\ref{sec:grpo_when}).
This complements \citet{kirichenko2025abstentionbench}, who find that reasoning
fine-tuning can \emph{degrade} calibrated abstention---together suggesting that
RL post-training is not a free improvement and should be applied where an
SFT-only model is demonstrably weak.

\paragraph{Reliability in data analytics.}
Statistical software (\eg, R, Stata) mechanically computes CIs and p-values
when explicitly requested, but does not decide which diagnostics to surface
or gate answers behind them.
Data quality detection systems
(Holoclean~\cite{rekatsinas2017holoclean}, RAHA~\cite{mahdavi2019raha})
flag errors \emph{in the data};
our system flags hazards \emph{in the query result} arising from query design
over clean data---a different failure mode.

% ============================================================
\section{Limitations}
\label{sec:limitations}

\paragraph{Profiler-visible statistics.}
Our model receives the profiler statistics block at inference time.
In a production deployment, a reliability profiler must run before the
model responds, adding query execution latency.
If the profiler is unavailable, reliability annotation degrades toward the
zero-shot baseline.

\paragraph{Taxonomy coverage.}
R1--R10 covers classical statistical hazards but does not capture
causal confounding, temporal autocorrelation, or measurement error.
The taxonomy is a starting point, not an exhaustive catalog.

\paragraph{Answer accuracy metric.}
Multi-row GROUP BY results are compared using exact row-set equality
after JSON serialization; without an ORDER BY clause, DuckDB's row ordering
is non-deterministic, which may undercount true matches.
This affects all models equally.

\paragraph{Enterprise schema generalization.}
Our Synthetic Customer evaluation schema is a 249-row synthetic table used for
schema analysis only; no private customer data appears in training or evaluation.
Whether models trained on public retail schemas generalize to proprietary
enterprise schemas requires further study.

\paragraph{Metric saturation and evaluation coverage.}
On our standard test sets the adequate-SFT model nears the reliability-F1
ceiling, which can mask differences between systems; we therefore also report a
strict exact-flag-set-match metric and a hard compound-hazard slice
($\geq 3$ flags). Two hazards are unevenly covered: R4 (missing-data bias) and
R8 (distribution-tail mismatch) depend on schema properties (null rates, skew)
and are absent from some schemas by construction---\eg, R4 cannot fire on a
table with no $>\!10\%$-null aggregable column---so per-flag conclusions for them
are schema-limited.

\paragraph{Scope of the out-of-distribution test.}
Our held-out H\&M domain probes a genuinely unseen schema and vocabulary, but its
single near-symmetric numeric column makes it somewhat \emph{easier} (R4/R8 do not
arise), so its absolute scores should be read as evidence of domain transfer
rather than of a harder benchmark. The in-/out-of-distribution GRPO comparison,
however, is controlled---same test set, both models---so the null GRPO effect
there is not an artifact of test difficulty.

% ============================================================
\section{Conclusion}

We introduced \sysname, a framework for statistical reliability annotation in
tabular QA, and used it to ask how much supervision the task actually requires.
SQL correctness and result reliability are independent: an executed result can
be factually accurate yet statistically meaningless. Our program-first data
pipeline avoids LLM-generation mode collapse by generating structural diversity
combinatorially. Our central empirical finding is that calibrated reliability
annotation is largely a \emph{data-efficiency} problem: a few hundred
schema-stratified SFT examples drive reliability-F1 to $0.98$ and \ucar\ to
zero, beat strong prompted baselines, and generalize to a held-out domain---
whereas GRPO with executable rewards, commonly assumed essential, helps only
when the SFT base is under-trained and adds nothing once it is adequate
(confirmed in-distribution, on a hard compound-hazard slice, under a strict
metric, and out-of-distribution). The practical takeaway for practitioners is
to invest a modest amount of additional supervised data before reaching for
reinforcement learning. We will release the pipeline code, reliability profiler,
and trained models upon acceptance to support further work on reliability-aware
data analytics.

% ============================================================
\section*{Ethics Statement}
The Synthetic Customer schema used for development is a synthetic table constructed
for analytical pipeline testing; no real customer data is included.
All training and evaluation uses publicly available retail datasets.

% ============================================================
\bibliographystyle{unsrtnat}
\bibliography{main}

% ============================================================
\appendix

\section{Program-First Data Generation}
\label{app:pipeline}

This appendix expands the summary in Section~\ref{sec:pipeline} with the query
grammar, the deterministic profiler rules, and a worked end-to-end example.
It is intended to make the pipeline reproducible.

\subsection{Why invert the paraphrase pipeline?}
\label{app:invert}
The conventional recipe for synthesizing analytics QA data is
\emph{paraphrase-first}: seed a set of natural-language questions, have an LLM
augment or paraphrase them for diversity, and only then derive the corresponding
SQL and any labels. Two problems follow. First, the \emph{distribution} of the
resulting queries is inherited from the LLM's generation prior, which collapses
to high-frequency templates (\eg, single-table counts, ``how many orders per
state?''); rare \emph{structural} hazards such as R3 (multiple-comparison
inflation) and R8 (distribution-tail mismatch) arise from query shapes the model
rarely proposes, so they are systematically under-represented. Second, the
reliability labels must be recovered \emph{after the fact} from text whose
executable meaning is no longer explicit.

We invert the dependency. The unit of generation is the \emph{program} (a SQL
query drawn from a grammar), not the sentence. Structural coverage is therefore
controlled by construction (Appendix~\ref{app:grammar}); each query is executed
and labeled deterministically, so the labels are ground truth rather than LLM
guesses (Appendix~\ref{app:profiler}); and the natural-language question is
\emph{rendered last}, as a surface realization of an item that is already fully
specified and labeled. NL diversity is then free to vary without perturbing the
hazard distribution.

\subsection{Layer 1: the query grammar}
\label{app:grammar}
Queries are sampled from the following grammar over a single flattened
per-customer/per-order table \texttt{T}:

\begin{verbatim}
query    ::= SELECT [gcols ","] agg FROM T
             [WHERE preds]
             [GROUP BY gcols
              [ORDER BY agg dir [LIMIT k]]]
agg      ::= AVG(c)  | MEDIAN(c) | SUM(c)
           | MAX(c)  | MIN(c)    | STDDEV(c)
           | P90(c)  | P10(c)    | COUNT(*)
           | COUNT(DISTINCT c)   | RATIO(c)
preds    ::= pred {AND pred}
pred     ::= c = v | c IN (v,...) | c > v
           | c < v | c BETWEEN v AND v
           | c IS [NOT] NULL | flag = {0,1}
gcols    ::= cat_col {"," cat_col}     (arity 0-2)
dir      ::= ASC | DESC
\end{verbatim}

\noindent
Each aggregate is restricted to type-compatible columns (\eg, \texttt{RATIO} only
applies to binary flag columns; \texttt{COUNT(DISTINCT)} only to categoricals).
The query is sampled along four structural dimensions whose marginals are fixed
by the weights in Table~\ref{tab:cfg-weights}; literal values are grounded in the
\emph{actual} column distributions (numeric thresholds sampled uniformly in the
empirical $[P_5, P_{95}]$ interval; categorical values drawn from the observed
domain, as an \texttt{IN}-list of three with probability $0.3$ and an equality
otherwise). Every candidate is executed in DuckDB and rejected if it returns zero
rows or a trivially constant result, guaranteeing that all retained queries are
answerable.

\begin{table}[H]
\centering
\footnotesize
\setlength{\tabcolsep}{4pt}
\begin{tabular}{lll}
\toprule
\textbf{Dimension} & \textbf{Values} & \textbf{Weights} \\
\midrule
Aggregate type   & 11 types        & uniform \\
Group-by arity   & 0/1/2           & .30/.50/.20 \\
Predicate count  & 0/1/2/3         & .20/.40/.30/.10 \\
Predicate kind   & cat/num/null/flag & .40/.25/.20/.15 \\
\bottomrule
\end{tabular}
\caption{Layer-1 sampling distribution over structural dimensions. If a group-by
  is present, an \texttt{ORDER BY} is added with probability $0.55$ (DESC with
  probability $0.70$), then a \texttt{LIMIT}\,$\in\!\{3,5,10\}$ with probability
  $0.50$. Coverage of rare hazards is a property of these marginals, not of an
  LLM's prior.}
\label{tab:cfg-weights}
\end{table}

\subsection{Layer 2: deterministic reliability profiler}
\label{app:profiler}
Each generated query is executed against its table, and the result set is passed
through a fixed rule per hazard---no model is involved, so labels are exact and
reproducible. Table~\ref{tab:profiler-rules} lists the firing condition for each
R1--R10. Confidence intervals use a Student-$t$ interval for means, $B{=}1000$
bootstrap resamples for other continuous aggregates, and a Wilson interval for
proportions.

\begin{table}[H]
\centering
\footnotesize
\begin{tabular}{ll}
\toprule
\textbf{Flag} & \textbf{Fires when} \\
\midrule
R1  & smallest group $n<5$ (abstain) or $<30$ (warn) \\
R2  & 95\% CI width $> 0.5\,\lvert\text{estimate}\rvert$ \\
R3  & runner-up's estimate lies within the \\
    & \quad winner's 95\% CI (top-$k$ not separable) \\
R4  & NULL rate of the aggregated column $>10\%$ \\
R5  & $\min/\max$ group size $<0.1$, or both $<10$ \\
R6  & deleting the winner's max row drops its \\
    & \quad mean below the runner-up (jack-knife) \\
R7  & overall leader loses its lead in $\geq1$ \\
    & \quad stratum of an unused covariate (Simpson) \\
R8  & $\lvert\text{skew}\rvert > 2$, or \\
    & \quad $\text{mean}/\text{median} > 1.5$ \\
R9  & proportion numerator $<30$ (Wilson) \\
R10 & expected cross-tab cell count $<5$ \\
\bottomrule
\end{tabular}
\caption{Deterministic firing conditions for the ten reliability hazards,
  as implemented in the profiler. Labels are multi-hot; a query may fire several.}
\label{tab:profiler-rules}
\end{table}

\subsection{Layers 3--4: balancing and NL realization}
\label{app:nl}
Layer~3 collapses queries to structural templates by abstracting their literals
and keeps one representative per template, then oversamples the naturally rare
labels (R3, R8) to a floor of $3{,}000$ instances each before realization.
Layer~4 renders two natural-language variants per query with a 3B
instruction-tuned model under one of five analyst personas, discarding any
variant whose embedding cosine similarity to an already-accepted question in the
batch exceeds $0.85$. Because the SQL, executed answer, and reliability labels
are fixed \emph{before} this step, NL diversity cannot distort the hazard
distribution.

\subsection{A worked example}
\label{app:example}
Consider one query sampled over the Olist schema (group-by arity 1, one
predicate, a top-$k$ ranking):

\noindent\begin{minipage}{\linewidth}
\begin{verbatim}
SELECT customer_state, AVG(review_score)
FROM orders
WHERE payment_type = 'voucher'
GROUP BY customer_state
ORDER BY AVG(review_score) DESC
LIMIT 5
\end{verbatim}
\end{minipage}

\noindent
Vouchers are a minority payment type, so once the data is grouped by state the
tail states hold very few orders. A representative execution returns a top state
with mean review $4.6$ over $n{=}14$ voucher orders, a runner-up at $4.5$ over
$n{=}22$, against a high-volume state (SP) with $n{\approx}900$. The profiler
then fires:
\begin{itemize}
  \item \textbf{R1 (warn)}: the smallest group in the result has $n{=}14<30$.
  \item \textbf{R3}: the $n{=}14$ winner's wide 95\% CI contains the runner-up's
    $4.5$, so the top-1 ranking is not statistically separable.
  \item \textbf{R5}: $\min/\max$ group size $=14/900\approx0.016<0.1$.
\end{itemize}
R4 does \emph{not} fire (\texttt{review\_score} is $<\!1\%$ null), nor does R8
(bounded 1--5 scores are not heavy-tailed), yielding the multi-hot label
\texttt{[R1, R3, R5]}. Layer~4 then realizes, \eg, ``Among orders paid with a
voucher, which five states have the highest average review score?'' The training
record pairs this question with the SQL, its executed answer, and the
reliability annotation. A paraphrase-first generator asked for ``diverse
analytics questions'' would rarely propose this rare-filter top-$k$ shape at all,
and would have no executable basis from which to derive the \texttt{[R1, R3, R5]}
labels.

\end{document}